\documentclass[11pt,a4paper]{article}
\usepackage[latin9]{inputenc}
\usepackage{float}
\usepackage{booktabs}
\usepackage{graphicx}
\usepackage[authoryear]{natbib}
\usepackage{subscript}
\usepackage[unicode=true]
 {hyperref}

\makeatletter

\pdfpageheight\paperheight
\pdfpagewidth\paperwidth

\providecommand{\tabularnewline}{\\}

\@ifundefined{date}{}{\date{}}

\usepackage[hyperref]{naaclhlt2019}
\usepackage{times}
\usepackage{latexsym}
\usepackage{url}

\aclfinalcopy 

\title{Amobee at SemEval-2019 Tasks 5 and 6: Multiple Choice CNN Over Contextual Embedding}

\author{Alon Rozental\thanks{~~These authors contributed equally to this work.}~~, Dadi Biton\footnotemark[1]~~\\
  Amobee Inc., Tel Aviv, Israel \\
   \tt alon.rozental@amobee.com \\ \tt dadi.biton@amobee.com }

\@ifundefined{showcaptionsetup}{}{%
 \PassOptionsToPackage{caption=false}{subfig}}
\usepackage{subfig}
\makeatother

\begin{document}
\maketitle 
\begin{abstract}
This article describes Amobee's participation in ``HatEval: Multilingual
detection of hate speech against immigrants and women in Twitter''
(task 5) and ``OffensEval: Identifying and Categorizing Offensive
Language in Social Media'' (task 6). The goal of task 5 was to detect
hate speech targeted to women and immigrants. The goal of task 6 was
to identify and categorized offensive language in social media, and
identify offense target. We present a novel type of convolutional
neural network called ``Multiple Choice CNN'' (MC-CNN) that we used
over our newly developed contextual embedding, \citet{MODBERT}\footnote{To be published.}.
For both tasks we used this architecture and achieved 4th place out
of 69 participants with an F\textsubscript{1} score of 0.53 in task
5, in task 6 achieved 2nd place (out of 75) in Sub-task B - automatic
categorization of offense types (our model reached places 18/2/7 out
of 103/75/65 for sub-tasks A, B and C respectively in task 6).
\end{abstract}

\section{Introduction}

\label{intro}
\begin{table*}[t]
\center\subfloat[]{%
\begin{tabular}{cc}
\toprule 
Sub-Task A & \tabularnewline
\midrule 
Label & Train\tabularnewline
\cmidrule[0.05em](rl){1-1}\cmidrule[0.05em](rl){2-2}Offensive & 4,400\tabularnewline
Not offensive & 8,840\tabularnewline
 & \tabularnewline
\midrule 
Total & 13,240\tabularnewline
\bottomrule
\end{tabular}}\subfloat[]{%
\begin{tabular}{cc}
\toprule 
Sub-Task B & \tabularnewline
\midrule 
Label & Train\tabularnewline
\cmidrule[0.05em](rl){1-1}\cmidrule[0.05em](rl){2-2}Targeted & 3,876\tabularnewline
Not targeted & 524\tabularnewline
 & \tabularnewline
\midrule 
Total & 4,400\tabularnewline
\bottomrule
\end{tabular}}\subfloat[]{%
\begin{tabular}{cc}
\toprule 
Sub-Task C & \tabularnewline
\midrule 
Label & Train\tabularnewline
\cmidrule[0.05em](rl){1-1}\cmidrule[0.05em](rl){2-2}Group & 1,074\tabularnewline
Individual & 2,407\tabularnewline
Other & 395\tabularnewline
\midrule 
Total & 3,876\tabularnewline
\bottomrule
\end{tabular}}\caption{\label{tab:Distributions-of-OffensEval}Distributions of lables in
OffensEval 2019.}
\end{table*}

Offensive language and hate speech identification are sub-fields of
natural language processing that explores the automatic inference
of offensive language and hate speech with its target from textual
data. The motivation to explore these sub-fields is to possibly limit
the hate speech and offensive language on user-generated content,
particularly, on social media. One popular social media platform for
researchers to study is Twitter, a social network website where people
``tweet'' short posts. Each post may contain URLs and/or mentions
of other entities on twitter. Among these ``tweets'' we can find
various opinions of people regarding political events, public figures,
products, etc. Hence, Twitter data turned into one of the main data
sources for both academia and industry. Its unique insights are relevant
for business intelligence, marketing and e-governance. This data also
benefits NLP tasks such as sentiment analysis, offensive language
detection, topic extraction, etc.

Both the OffensEval 2019 \href{https://competitions.codalab.org/competitions/20011}{task}
(\citet{offenseval}) and HatEval 2019 \href{https://competitions.codalab.org/competitions/19935\#learn_the_details}{task}
are part of the \href{http://alt.qcri.org/semeval2019/}{SemEval-2019}
workshop. OffensEval has 3 sub-tasks with over 65 groups who participate
in each sub-task and HatEval has 2 sub-tasks with 69 groups.

Word embedding is one of the most popular representations of document
vocabulary in low-dimensional vector. It is capable of capturing context
of a word in a document, semantic and syntactic similarity, relation
with other words, etc. For this work, word embedding was created with
a model similar to Bidirectional Encoder Representations from Transformers
(BERT), \citet{devlin2018bert}. BERT is a language representation
model designed to pre-train deep bidirectional representations by
jointly conditioning on both left and right context in all layers.
As a result, the pre-trained BERT representations can be fine-tuned
to create state-of-the-art models for a wide range of tasks, such
as question answering and language inference, without substantial
task-specific architecture modifications. Besides the word embedding,
BERT generates a classification token, which can be used for text
classification tasks. 

This paper describes our system for the OffensEval 2019 and HatEval
2019 tasks, where our new contribution is the use of contextual embedding
(modified BERT) together with an appropriate network architecture
for such embeddings .

The paper is organized as follows: Section 2 describes the datasets
we used and the pre-process phase. Section 3 describes our system
architecture and presents the MC-CNN. In section 4 we present the
results of both tasks - the OffensEval and HatEval. Finally, in section
5 we review and conclude the system.

\section{Data and Pre-Processing}

\begin{figure*}[t]
\includegraphics[scale=0.45]{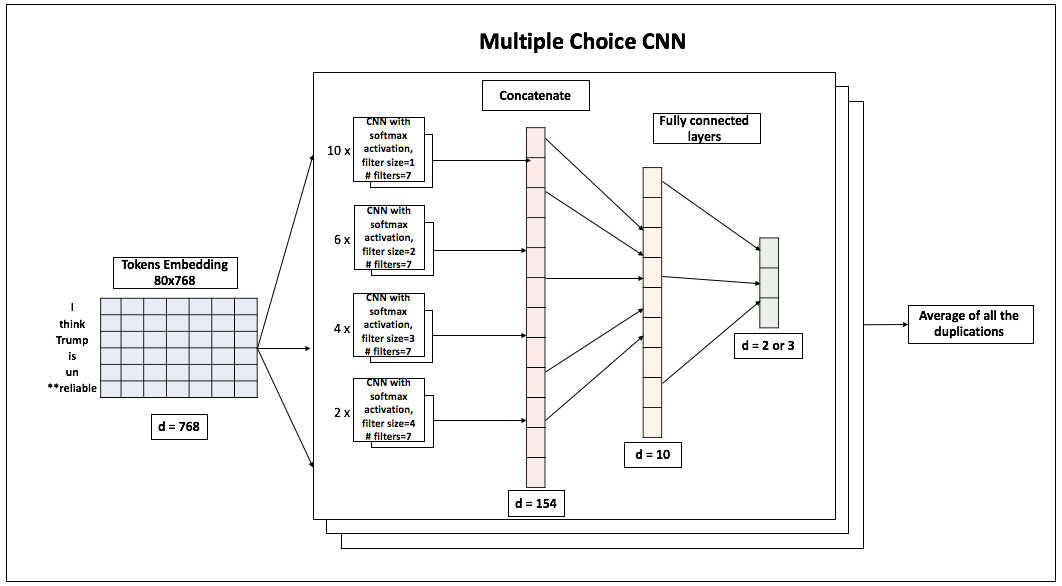}\caption{\label{fig:Architecture-of-Amobee-1}Architecture of Amobee offensive
language detector.}
\end{figure*}
We used Twitter Firehose dataset. We took a random sample of 50 million
unique tweets using the Twitter Firehose service. The tweets were
used to train language models and word embeddings; in the following,
we will refer to this as the Tweets\_50M dataset.

A modified language model, based on BERT, was trained using a large
Tweets\_50M dataset, containing 50 million unique tweets. We trained
two models, one used to predict hate speech in posts (task 5) and
the other used to predict offensive language in posts (task 6). The
pre-process on the Tweets\_50M dataset consists of replacing URLs
and Twitter handles with special tokens and keeping only the first
80 sub-word tokens in each tweet (for our vocabulary over 99\% of
the tweets contain less than 80 tokens).

The language model we trained differs from \citet{devlin2018bert}
mainly by the addition of a latent variable that represents the topic
of the tweet and the persona of the writer. The work on this model
is still in progress and in this work we have used an early version
of the model described in \citet{MODBERT}.

\subsection{OffensEval}

OffensEval 2019 is divided into three sub-tasks.
\begin{enumerate}
\item Sub-task A - Offensive language identification - identify whether
a text contains any form of non-acceptable language (profanity) or
a targeted offense.
\item Sub-task B - Automatic categorization of offense types - identify
whether a text contains targeted or non-targeted profanity and swearing.
\item Sub-task C - Offense target identification - determine whether the
target of the offensive text is an individual, group or other (e.g.,
an organization, a situation, an event, or an issue).
\end{enumerate}
The official OffensEval task datasets, retrieved from social media
(Twitter). Table \ref{tab:Distributions-of-OffensEval} presents the
label distribution for each sub-task. For further shared task description,
data-sets and results regarding this task, see \citet{zampierietal2019}.

\subsection{HatEval}

HatEval 2019 is divided into two sub-tasks.
\begin{enumerate}
\item Sub-task A - Hate Speech Detection against Immigrants and Women: a
two-class classification where systems have to predict whether a tweet
in English with a given target (women or immigrants) is hateful or
not hateful.
\item Sub-task B - Aggressive behavior and Target Classification: where
systems are asked first to classify hateful tweets for English and
Spanish (e.g., tweets where Hate Speech against women or immigrants
has been identified) as aggressive or not aggressive, and second to
identify the target harassed as individual or generic (i.e. single
human or group). In this paper we will focus only on sub-task A as
none of the participants overcame the baseline accuracy in sub-task
B.
\end{enumerate}
There were 69 groups who participated in sub-task A. Table \ref{tab:Distributions-of-HatEval}
presents the label distribution in sub-task A. For further shared
task description, data-sets and results regarding this task, see \citet{hateval2019semeval}.
HatEval also included Spanish task which we didn't participate in.
\begin{table}
\center%
\begin{tabular}{cc}
\toprule 
Label & Train\tabularnewline
\cmidrule[0.05em](rl){1-1}\cmidrule[0.05em](rl){2-2}Hate speech & 4,210\tabularnewline
Not hate speech & 5,790\tabularnewline
 & \tabularnewline
\midrule 
Total & 10,000\tabularnewline
\bottomrule
\end{tabular}\caption{\label{tab:Distributions-of-HatEval}Distributions of lables in HatEval
2019.}
\end{table}

\section{Multiple Choice CNN}

For both tasks, using our contextual word embedding, we tried several
basic approaches - A feed forward network using the classification
vector and an LSTM and simple CNNs \citet{zhang2015sensitivity} using
the words vectors. These approaches overfitted very fast, even for
straightforward unigram CNN with 1 filter, and their results were
inferior to those obtained by similar models over a Twitter specific,
Word2Vec based embedding, \citet{mikolov2013distributed,rozental2018amobee}.
The fast overfitting is due to the information contained in contextual
embedding which was not reflected in Word2Vec based embedding.

In order to avoid overfitting and achieve better results we created
the MC-CNN model. The motivation behind this model is to replace quantitative
questions such as ``how mad is the speaker?'', where the result
is believed to be represented by the activation of the corresponding
filter, with multiple choice questions such as ``what is the speaker
- happy/sad/other?'', where the number of choices denoted by the
number of filters. By forcing the sum of the filter activations for
each group to be equal to 1, we believe that we have acheived this
effect.

The model that produced the best results is an ensemble of multiple
MC-CNNs over our developed contextual embedding, described in figure
\ref{fig:Architecture-of-Amobee-1}. On top of our contextual embedding,
we used four filter sizes - 1-4 sub-word token n-grams. For each filter
size individual filters were divided into groups of 7 and a softmax
activation applied on the output of each group. These outputs were
concatenated and passed to a fully connected feed forward layer of
size 10 with tanh activation before it yeiled the networks' prediction.
To decrease the variance of the results, there were multiple duplications
of this architecture, where the final prediction was the average of
all the duplications' output.

\section{Results}

\label{sec:results}

We chose to use this architecture for both tasks because we believe
that the BERT model output contains most of the information about
the tweet. The layers above, the MC-CNN and the fully connected layers,
adapt it to the given task. We think that this model can be use for
variety of NLP tasks in twitter data with the appropriate hyper-parameters
tuning.

The results yielded from the architecture which was described in figure
\ref{fig:Architecture-of-Amobee-1} for both tasks. We optimized the
hyper-parameters to maximize the F\textsubscript{1}score using categorical
cross entropy loss. The tuned parameters were the activation function
of the filters and the number of filters in the MC-CNNs, the size
of the filter groups of the MC-CNN, and the hidden layer size. The
best result were achieved with a sigmoid activation function on the
filters, where the number of filters was 7 in each group. There were
10, 6, 4 and 2 filter groups for filter sizes of 1, 2, 3 and 4 respectively.
The model with those hyper-parameters yielded the best results in
both tasks.

At HatEval the model achieved an F\textsubscript{1}score of 0.535.
In table \ref{tab:Results-HatEval} there is the best result compared
to two baselines- linear Support Vector Machine based on a TF-IDF
representation (SVC), and a trivial model that assigns the most frequent
label (MFC), according to the F\textsubscript{1}score.
\begin{table}[H]
\begin{tabular}{|lll|}
\hline 
\textbf{System} & \textbf{F1 (macro)} & \textbf{Accuracy}\tabularnewline
\hline 
SVC baseline & 0.451 & 0.492\tabularnewline
MFC baseline & 0.367 & 0.579\tabularnewline
\hline 
\textbf{MC-CNN} & \textbf{0.535} & \textbf{0.558}\tabularnewline
\hline 
\end{tabular}\caption{\label{tab:Results-HatEval}F\protect\textsubscript{1}Score and Accuracy
of MC-CNN Comparing to Baselines at HatEval.}
\end{table}

At OffensEval the model achieved an F\textsubscript{1}score of 0.787,
0.739 and 0.591 for sub-tasks A, B and C respectively. In table \ref{tab:FScore-and-Accuracy}
there is the best result compared to the baseline for sub-tasks A,
B and C respectively according to the F\textsubscript{1}score.
\begin{table}
\subfloat[\label{tab:Results-a}Sub-task A.]{%
\begin{tabular}{|lll|}
\hline 
\textbf{System} & \textbf{F1 (macro)} & \textbf{Accuracy}\tabularnewline
\hline 
All NOT baseline & 0.4189 & 0.7209\tabularnewline
All OFF baseline & 0.2182 & 0.2790\tabularnewline
\hline 
\textbf{MC-CNN} & \textbf{0.7868} & \textbf{0.8384}\tabularnewline
\hline 
\end{tabular}}

\subfloat[\label{tab:Results-b}Sub-task B.]{%
\begin{tabular}{|lll|}
\hline 
\textbf{System} & \textbf{F1 (macro)} & \textbf{Accuracy}\tabularnewline
\hline 
All TIN baseline & 0.4702 & 0.8875\tabularnewline
All UNT baseline & 0.1011 & 0.1125\tabularnewline
\hline 
\textbf{MC-CNN} & \textbf{0.7386} & \textbf{0.9042}\tabularnewline
\hline 
\end{tabular}}

\subfloat[\label{tab:Results-c}Sub-task C.]{%
\begin{tabular}{|lll|}
\hline 
\textbf{System} & \textbf{F1 (macro)} & \textbf{Accuracy}\tabularnewline
\hline 
All GRP baseline & 0.1787 & 0.3662\tabularnewline
All IND baseline & 0.2130 & 0.4695\tabularnewline
All OTH baseline & 0.0941 & 0.1643\tabularnewline
\hline 
\textbf{MC-CNN} & \textbf{0.5909} & \textbf{0.7042}\tabularnewline
\hline 
\end{tabular}}

\caption{\label{tab:FScore-and-Accuracy}F\protect\textsubscript{1}Score and
Accuracy of MC-CNN Comparing to Baselines at OffensEval.}
\end{table}

\section{Conclusion}

In this paper we described the system Amobee developed for the HatEval
and OffensEval tasks. It consists of our novel task specific contextual
embedding and MC-CNNs with softmax activation. The use of social networks
motivated us to train contextual embedding based on the Twitter dataset,
and use the information learned in this language model to identify
offensive language and hate speech in the text. Using MC-CNN helped
overcome the overfitting caused by the embedding. In order to decrease
the variance of the system we used duplications of this model and
averaged the results. This system reached 4th place at the HateEval
task with an F\textsubscript{1}score of 0.535, and 2nd place at sub-task
B in the OffensEval task, with an F\textsubscript{1}score of 0.739.
As we mentioned, we used an early version of a Twitter specific language
model to achieve the above results. We plan to release the complete,
fully trained version in the near future and test it for different
NLP tasks- such as topic classification, sentiment analysis, etc.

\bibliographystyle{acl_natbib}
\bibliography{semeval}

\end{document}